\documentclass[]{article}

\usepackage{amssymb}
\usepackage{amsmath}

\usepackage{setspace}

\usepackage{fullpage}

\newcommand{\tuple}[1]{\vec{#1}}
\newcommand {\indep}[3] {#2 ~\bot_{#1}~ #3}
\newcommand {\indepc}[2] {#1 ~\bot~ #2}

\def\dom{\mbox{Dom}}

\def\part{\mathcal{P}}

\def\bel{\mbox{Bel}}

\def\believes{B}
\def\possible{P}
\def\doubted{\mbox{\textexclamdown}}
\def\doubting{\mbox{!}}
\def\disbelieve{D}
\def\regardless{R}
\def\confimp{\xrightarrow{\oplus}}
\def\credimp{\xrightarrow{\otimes}}
\def\skepimp{\xrightarrow{\ominus}}
\def\openimp{\xrightarrow{\odot}}
\def\var{\mbox{Var}}

\begin{document}
\title{The Doxastic Interpretation of Team Semantics}
\author{Pietro Galliani\\University of Helsinki\\Finland\\(pgallian{@}gmail.com)\thanks{This work was supported by the European Science Foundation Eurocores programme
LogICCC [FP002 - Logic for Interaction (LINT)], by the V\"ais\"ala Foundation and by grant 264917 of the Academy of Finland.}}
\maketitle
\begin{abstract}
We advance a \emph{doxastic interpretation} for many of the logical connectives considered in Dependence Logic and in its extensions, and we argue that Team Semantics is a natural framework for reasoning about beliefs and belief updates.
\end{abstract}
\section{Introduction}
Dependence Logic \cite{vaananen07} is an extension of First Order Logic which adds to its language \emph{dependence atoms} of the form $=\!\!(\tuple t_1, \tuple t_2)$, where $\tuple t_1$ and $\tuple t_2$ are tuples of terms\footnote{Sometimes only atoms of the form $=\!\!(\tuple t, t')$ are taken as primitive, where $t'$ is a single term. $=\!\!(\tuple t_1, \tuple t_2)$ is then equivalent to $\bigwedge_{t' \in \tuple t_2} =\!\!(\tuple t, t')$.}, with the intended interpretation of ``the value of $\tuple t_2$ is a function of the value of $\tuple t_1$.'' It is a first-order \emph{logic of imperfect information}, like IF Logic \cite{hintikkasandu89,hintikka96,mann11} or Branching Quantifier Logic \cite{henkin61}; but rather than adding new possible patterns of dependence or independence between \emph{quantifiers}, as these logics do, Dependence Logic isolates the notion of dependence away from the one of quantification and permits the examination of patterns of dependence and independence between \emph{variables} or, more in general, between tuples of terms.

This different outlook makes Dependence Logic a most suitable framework for the formal study, in a first-order setting, of functional dependence itself; and, furthermore, this logic is readily adaptable to the analysis of other, non-functional notions of dependence or independence \cite{gradel13,engstrom12a,galliani12}.

Like other logics of imperfect information, Dependence Logic admits both a Game Theoretic Semantics, an imperfect information variant of the one for First Order Logic, and a \emph{Team Semantics}, a compositional semantics which is a natural adaptation of Hodges' Trump Semantics \cite{hodges97}. One striking peculiarity of the current state of the art of the research in Dependence Logic and its extensions is a willingness to take Team Semantics -- and not Game Theoretic Semantics, as for the case of much IF Logic research -- as the fundamental semantic framework; and this different approach is at the root of many recent technical developments in the field, such as, for example, the characterizations of team class definability of \cite{kontinenv09}, \cite{kontinennu09} and \cite{galliani12}, the hierarchy results of \cite{durand11}, and the study of notions of generalized quantification of \cite{engstrom12a} and \cite{engstrom12}.

This paper is a detailed account of a \emph{doxastic} interpretation for Team Semantics, according to which formulas are to be interpreted as assertions about beliefs and belief updates. This is not a novel idea: as a matter of fact, it is already implicit in the equivalence proof between Trump Semantics and Game Theoretic Semantics of \cite{hodges97}. However, the consequences of this insight are far from fully explored: until now, doxastic concerns have played very little role in the development of extensions of Dependence Logic, and the doxastic meanings of known connectives have been left largely unexamined.\footnote{Even such important connectives as the Dependence Logic disjunction, which -- after \cite{vaananen07b} -- we will write $\phi \otimes \psi$, or the \emph{intuitionistic implication} $\phi \rightarrow \psi$ of \cite{abramsky09} do not have, to the knowledge of the author, any known doxastic interpretation. Later in this work, we will provide such interpretations: as we will see, these connectives can be understood in terms of \emph{belief updates}.}

In what follows, we will gradually develop a formal system -- in essence, a notational variant of Jouko V\"a\"an\"anen's Team Logic \cite{vaananen07b} -- and show that many of the atoms, connectives and operators of Team Semantics arise naturally from concerns about beliefs and belief updates. It is the hope of the author that by doing this, we may succeed in drawing attention to this interesting avenue of research. 
\section{Belief Models}
\label{sect:belief_models}
Let $M$ be a first order model with at least two elements in its domain, let $V \subseteq \var$ be a set of \emph{state variables}, and let us consider the set of all first-order \emph{assignments} over $\dom(M)$ with domain $V$.

With respect to Tarski's semantics for first order logic, such an assignment $s$ represents a possible \emph{state of things}: in other words, once the model $M$ is fixed, the truth value of a first order formula in an assignment depends on the elements of the model that the assignment associates to all the free variables of the formula, \emph{and on nothing else}.

Even disregarding First Order Logic, first order assignments are very natural objects for representing states of things. For example, let us suppose that our model's domain contains all the participants to a given contest, and that our states represent the possible outcomes of the contest -- and, in particular, the players who obtained the first three positions in the final ranking.

Such an outcome can be represented, in an obvious way, as an assignment $s$ over $M$ with domain $\{w_1, w_2, w_3\}$: here, $s(w_1)$ would be the identity of the winner, and so on.

For this particular example, of course, we would need to add a constraint requiring that no one can be placed in two different positions in the final ranking: this can be represented easily enough as the first order axiom
\begin{equation}
	\phi := \lnot \exists x ((w_1 = x \wedge w_2 = x) \vee (w_2 = x \wedge w_3 = x) \vee (w_1 = x \wedge w_3 = x))
\label{eqn:axiom_one_pos}
\end{equation}
as a condition that must hold for \emph{all} possible states of things $s$, in the sense that $s$ is an acceptable outcome if and only if $M \models_s \phi$. 

Once we have added this axiom, there is not much left to do: as long as the domain of the model is the set of all participants to the contest, any assignment which satisfies the above formula represents a possible contest outcome. 

A special case of this which is of no small interest is when the domain of the model $M$ consists of only two elements $0$ and $1$, or ``False'' and ``True'': then an assignment is easily seen to be equivalent to a propositional truth valuation, or, if one prefers, to the specification to a \emph{possible world} in the sense of Kripke's Semantics for Modal Logic.

Now, let us return to our example, and let us consider an agent $A$ who has some -- not necessarily true, nor complete -- belief about who will reach the first three places of our tournament. How can we represent this belief?

There are many possible choices here: for example, we could consider a probability distribution over states, or a possibility distribution \cite{zadeh78}, or even a Dempster-Shafer distribution \cite{shafer76,dempster08}.

But let us limit ourselves to a very simple idea, and consider the set $X_A \subseteq (V \mapsto \dom(M))$ of all possible states of things (assignments) which our agent believes to be possible. This idea of representing beliefs as ``sets of possible states'' is fairly common in knowledge representation theory, and -- even though other approaches, such as the ones described above, are certainly more sophisticated -- it is a reasonable starting point.

Furthermore, this approach plays on the analogy between our framework and modal logic: indeed, it is easy to see that, at least for the case of Boolean models, such a belief set is exactly a set of \emph{possible worlds} which an agent can see from the ``actual'' world. An important difference between our framework and modal logic, however, is that in our case the agents can reason only about \emph{outcomes}, and not about their beliefs or about the beliefs of other agents. 

Now, what can we do with beliefs? To begin with, we can describe their properties in some suitable logical formalism; and, as Section \ref{sect:atoms}. will show, many primitive formulas considered in logics of imperfect information have a very natural interpretation in these terms.

But we can also \emph{update} beliefs. In general, a (unary) \emph{update operation} will be a function $O$ from belief sets, or, to use the terminology in common use for logics of imperfect information, from \emph{teams}, to sets of belief sets. We will not require these updates to be deterministic; and we will write $O(X) \mapsto Y$ as a shorthand for $Y \in O(X)$, that is, for the statement that $Y$ is a possible outcome of updating $X$ according to the rule $O$.

\emph{Binary} update operators are defined analogously, as functions $\diamond$ mapping each pair of teams $X$ and $Y$ to a set $X \diamond Y$ of possible resulting teams; and again, we will write $X \diamond Y \mapsto Z$ for $Z \in (X \diamond Y)$, that is, for stating that the belief set $Z$ is a possible outcome of updating $X$ with $Y$ according to the rule $\diamond$.

Ternary or $n$-ary operators can also be defined in the same way, but we will not need to consider any of them in the present work.

Once we have belief operators and a language for describing properties of belief sets we can ask a number of new questions, such as
\begin{enumerate}
\item Can a certain belief set be seen as the result of a certain update being applied between belief sets satisfying certain properties?
\item If we update a belief set under a certain rule, and the other belief sets used for the update (if any) satisfy certain properties, can we guarantee that the resulting belief set will satisfy certain other properties?
\end{enumerate}
This kind of question is of clear practical importance in Artificial Intelligence: if some intelligent system's belief state respects a condition $\phi$ and our system interacts with some other system whose belief state respects another condition $\psi$, can we guarantee that the resulting belief states will respect some further condition $\theta$?

The whole discipline of \emph{belief revision} \cite{gardenfors92}, for example, can be understood as a special case of this, as its fundamental problem consists in the study and comparison of the ways of updating a knowledge base $K$ when a new statement $\phi$ is learned which is contradictory to it.

Our framework, in itself, is vastly more general -- and, of course, vastly more computationally expensive -- than any system of belief revision; but on the other hand, the update operations that we will discuss here are all much simpler than those considered in Belief Revision. What, in the opinion of the author, logics of imperfect information can provide to that field is a very general logical framework for defining update operations and reasoning about their properties.

But enough chatter. Beginning with the next section, we will define primitives and operators for a very general logic of imperfect information, containing most of the connectives which have been studied so far in the context of logics of imperfect information, and we will discuss their doxastic interpretations.
\section{Atoms and First Order Formulas}
\label{sect:atoms}
In this section, we will gradually develop a logical formalism -- basically, a fragment of Team Logic -- and use it to state properties of belief sets.

Now, what can be said about the beliefs of an agent $A$, or, to be more precise, about his belief set $X_A$? 

To begin with, we can ask, given a first order\footnote{We could also use here any extension or variant of First Order Logic which admits a Tarski-style semantics, such as Transitive Closure Logic or First Order Logic augmented with the H\"artig quantifier.} formula $\phi$, whether our agent $A$ \emph{believes} that $\phi$ holds. This justifies the following semantic rule:
\begin{description}
	\item[DI-bel] If $\phi$ is first order, $M \models_X \believes(\phi)$ if and only if $M \models_s \phi$ for all $s \in X$ 
\end{description}
where the expression $M \models_s \phi$ means that the assignment $s$ satisfies $\phi$ in $M$ according to the usual Tarski semantics. 

As an example, let us consider again the scenario described in the previous section and the formula $\phi$ of Equation (\ref{eqn:axiom_one_pos}). Then ``$M \models_{X_A} \believes(\phi)$'' is a sanity condition for our agent, corresponding to the statement according to which he believes that no player will get two distinct positions in the final rankings.

As another, perhaps quite unnecessary, example, suppose that our agent $A$ believes that the winner of the contest will be female; then, for all $s \in X_A$ we will have that $M \models_s \mbox{Female}(w_1)$, and hence that $M \models_{X_A} \believes(\mbox{Female}(w_1))$.

In most logics of imperfect information, one would just write $M \models_X \phi$ for what we would write here as $M \models_X \believes(\phi)$. Furthermore, the above condition would be given just for first-order literals, and we would rely on the connectives of our logic in order to build expressions equivalent to $\believes(\phi)$ for complex first-order formulas $\phi$: indeed, as mentioned in (\cite{vaananen07}, Proposition 3.30), if $\phi$ is first order then $M \models_X \phi$ if and only if $M \models_s \phi$ for all $s \in X$.

Here, however, we will write the $\believes$ operator explicitly, for three different reasons. First of all, our objective is emphatically \emph{not} to develop a terse formalism in which to express everything that can be expressed in a logic of imperfect information: instead, we want to illustrate a possible interpretation of logics of imperfect information, and hence it will be useful to examine many doxastically significant conditions and operators. Furthermore, it is vital for our purposes to distinguish between the first-order level of our language, which allows us to summarize the properties of all the assignments of the team \emph{taken individually}, the more sophisticated kinds of atoms that we will describe later in this section, and the update connectives that we will introduce in Sections \ref{sect:updates}-\ref{sect:quantifiers}. The fact that first-order formulas can be decomposed in terms of first-order literals and update operators will then be, from this point of view, an interesting \emph{theorem}, not something built in our definitions. Finally, and perhaps more practically, we want to be able to express another kind of ``first-order'' assertion in our language, and we need to distinguish it from belief statements as those just considered.

This new kind of first-order assertion is a \emph{possibility} assertion, which corresponds to our agent believing some first order condition to be \emph{possibly the case} in the ``true'' assignment. We can introduce this kind of assertion as follows:
\begin{description}
	\item[DI-pos:] If $\phi$ is first order, $M \models_X \possible(\phi)$ if and only if $M \models_s \phi$ for some $s \in X$ 
\end{description}
Dependence Logic and IF-Logic, the two most studied logics of imperfect information, are downwards closed\footnote{A logic of imperfect information is said to be \emph{downwards closed} if for all models $M$, teams $X$ and $Y$ and expressions $\phi$, $M \models_X \phi$ and $Y \subseteq X$ implies $M \models_Y \phi$.}  and hence incapable of expressing this sort of statement. The most widely known formalism capable of that is Team Logic \cite{vaananen07b}, where $\possible(\phi)$ corresponds to $\sim \lnot \phi$, where $\lnot$ is the dual negation and $\sim$ is the contradictory one. 

From the point of view of the present work, possibility statements are very natural: for example, using them we can express that our agent $A$ considers it \emph{possible} that the winner will be female, that is, that $M \models_{X_A}\possible(\mbox{Female}(w_1))$.

We could also give and justify along similar lines further ``first-order'' conditions over belief sets: for example, we could state that an expression of the form $\mbox{Most}(\phi)$ holds in a team $X$ if and only if most of the assignments in $X$ satisfy $\phi$. But let us now move to conditions which cannot be verified by examining the truth value of a first order formula in all assignments of our belief set.

In the example which we are considering, what else could our agent $A$ believe that we cannot express already? Well, to begin with, our agent could believe that he knows the identity of the winner.

It is easy enough to assert, using what we already have, that the agent believes that the winner will be $a_0$ for some $a_0 \in \dom(M)$: indeed, this is precisely the condition corresponding to $M \models_{X_A} \believes(w_1 = a_0)$. But what we are asking now is different: we want a formula that specifies that the agent believes that \emph{someone} will certainly be the winner, but does not specify who this person is.

An existential quantifier will not do the trick: writing $M \models_{X_A} \believes(\exists x (w_1 = x))$ corresponds only to asserting that our agent believes that someone will win the contest, not that he believes this about someone in particular!\footnote{In brief, we are discussing here the \emph{de dicto}/\emph{de re} distinction.} For example, if Tom, Bob and Jack are three participants to the tournament, it is easy to see that the belief set
\[
	X_A = \begin{array}{c | c c c}
		& w_1 & w_2 & w_3\\
		\hline
		s_0 & \mbox{Tom} & \mbox{Bob} & \mbox{Jack}\\
		s_1 & \mbox{Bob} & \mbox{Tom} & \mbox{Jack}
	\end{array}
\]
corresponding to the belief state in which our agent $A$ knows that Jack will get third place, but is unsure about who between Tom and Bob will get the second place and who the first one, satisfies the formula $\believes(\exists x (w_1 = x))$ but it does not respect the condition we are talking about.

In fact, \emph{no} expression of the form $\believes(\phi)$ or $\possible(\phi)$ will allow us to express our intended condition: indeed, those of the former sort are \emph{flat} in the sense of \cite{vaananen07}, and hence hold in a team if and only if they hold in all singleton subteams, and those of the latter one are \emph{upwards closed}, in the sense that if $M \models_X \possible(\phi)$ and $X \subseteq Y$ then $M \models_Y \possible(\phi)$. 

What we seem to need is some way of saying that the value of $w_1$ is \emph{the same} for all assignments in our team. This is precisely the semantics for \emph{constancy atoms} of Dependence Logic.\footnote{Another reasonable approach could be to define an ``external'' existential quantifier $\exists$, and model the intended condition as $\exists x \believes(x = w_1)$. Later in this section, we will briefly explore this idea; but we can anticipate that this external existential quantifier is precisely the $\exists^1$ operator of \cite{kontinenv09}.} 
\begin{description}
	\item[DI-con:] For all terms $t$, $M \models_X =\!\!(t)$ if and only if, for all $s, s' \in X$, $t\langle s\rangle = t\langle s'\rangle$. 
\end{description}
Given this definition, $=\!\!(w_1)$ characterizes precisely the condition of our example; and, more in general, it is easy to see that $=\!\!(t)$ is satisfied by a team $X_A$ if and only if the corresponding agent $A$ believes that he knows the value of $t$.

What if our agent instead believes that he would be able to guess the name of the winner from the name of the second placed participant? This corresponds nicely to the \emph{dependence atom} $=\!\!(w_2, w_1)$, where the rule for dependence atoms is
\begin{description}
	\item[DI-dep:] $M \models_X =\!\!(\tuple t_1, \tuple t_2)$ if and only if any two $s, s' \in X$ which assign the same value to $\tuple t_1$ also assign the same value to $\tuple t_2$.
\end{description}

What else can we say about our agent $A$'s beliefs? For example, he might think that everybody who has a chance to make it to first place has also a chance to make it to second place: then, for all $s \in X_A$  there exists a $s' \in X_A$ such that $s(w_1) = s'(w_2)$.  This is represented by the \emph{inclusion atom} $w_1 \subseteq w_2$ from \cite{galliani12}, where
\begin{description}
	\item[DI-inc:] For all tuples of terms $\tuple t_1$ and $\tuple t_2$, of the same length, $M \models_X \tuple t_1 \subseteq \tuple t_2$ if and only if for every $s \in X$ there exists a $s' \in X$ such that $\tuple t_1\langle s\rangle = \tuple t_2\langle s'\rangle$.
\end{description}
The meaning of the expression $w_1 \subseteq w_2$ is of course different from that of $\believes(w_1 = w_2)$, which would instead state that the agent believes that the first and second placed players will be the same.

As an aside, possibility operators can also be constructed in terms of inclusion atoms: indeed, if the free variables of $\phi$ are $x_1 \ldots x_n$ then it is easy to see that $\possible(\phi)$ can be written as $\exists w_1 \ldots w_n (\phi[w_1 \ldots w_n/x_1 \ldots x_n] \wedge (w_1 \ldots w_n \subseteq x_1 \ldots x_n))$. 

Again, what else? Well, our agent could also think that no one who has some chance to take first place has also some chance to take third place -- only first or second. This is represented by the \emph{exclusion atom} $w_1 ~|~ w_3$, also from \cite{galliani12}, where 
\begin{description}
	\item[DI-exc:] For all tuples of terms $\tuple t_1$ and $\tuple t_2$ of the same length, $M \models_X \tuple t_1 ~|~ \tuple t_2$ if and only if for all $s, s' \in X$, $\tuple t_1\langle s\rangle \not = \tuple t_2\langle s'\rangle$.
\end{description}
Of course, this is different, and stronger, than $\believes(w_1 \not = w_3)$, which would only state that the agent believes that the winner and the third placed player will not be the same. 

For the last pair of atoms that we will describe here, we need to modify slightly our example. Let us suppose that $w_1$, $w_2$ and $w_3$ represent the winners of three \emph{different} tournaments in three successive years, so that the same player could conceivably win more than one of them. Then a possible situation might be that, in the opinion of the agent, learning the winner of the first year would not tell him anything about who the winner of the third year that he does not know already -- or, in other words, that the set of all possible third year winners is the same for each fixed first year winner. This corresponds to the \emph{independence atoms} $\indepc{w_1}{w_3}$, where
\begin{description}
	\item[DI-ind$_C$:] Let $\tuple t_1$ and $\tuple t_2$ be two tuples of terms, not necessarily of the same length. Then $M \models_X \indepc{\tuple t_1}{\tuple t_2}$ if and only if for all $s, s' \in X$ there exists a $s'' \in X$ with $\tuple t_1\langle s''\rangle  = \tuple t_1 \langle s\rangle $ and $\tuple t_2\langle s''\rangle = \tuple t_2 \langle s'\rangle$.
\end{description}
As pointed out in \cite{gradel13}, $\indepc{t}{t}$ is logically equivalent to $=\!\!(t)$. The reason for this is clear: in our interpretation, $\indepc{t}{t}$ means that our agent thinks that he would not learn anything new about $t$ by being told the value of $t$, and this is possible only if he believes that he knows it already.

Finally, our agent may think that, for the purpose of learning who will be the winner in the third year, learning who won in the first and second year and knowing who won in the second year alone makes no difference. Perhaps, according to this agent, learning who won in the second year \emph{would} give him valuable information, and so would learning who won in the first year; but once he learned who won in the second year, learning who won in the first year too would be quite irrelevant.

As an example of this situation, let us consider the following belief set: 
\[
	X_A = \begin{array}{c | c c c}
		& w_1 & w_2 & w_3\\
		\hline
		s_0 & \mbox{Bob} & \mbox{Tom} & \mbox{Tom}\\
		s_1 & \mbox{Tom} & \mbox{Bob} & \mbox{Bob}\\
		s_2 & \mbox{Tom} & \mbox{Bob} & \mbox{Jack}\\
		s_3 & \mbox{Jack} & \mbox{Bob} & \mbox{Bob}\\
		s_4 & \mbox{Jack} & \mbox{Bob} & \mbox{Jack}\\

	\end{array}
\]
Here our agent believes that if Tom won the second year  then he will win the third year too, and that if Bob won the second year then either him or Jack will win the third year. So, learning who won the second year would allow him to infer something about who will win the third year. Also, he believes that if Tom won the first year then one of  Bob or Jack will win the third year, that if Jack won the first year then one of  Bob or Jack will win the third year and that if Bob won the first year then Tom will certainly win the third year. So, learning who won the first year would allow him to infer something about who will win the third year. However, suppose that Tom is told who won the second year. Then learning who won in the first year too would tell him nothing new about who will win the third year: if the winner of the second year is Tom, then he is also the winner of the third year, and if the winner of the second year is Bob then, no matter who won the first year, both Bob and Jack are possible winners for the third year competition.

This is modeled by the following, more general \emph{independence atom}, also from \cite{gradel13}:
\begin{description}
	\item[DI-ind:] Let $\tuple t_1$, $\tuple t_2$ and $\tuple t_3$ be two tuples of terms, not necessarily of the same length. Then $M \models_X \indep{\tuple t_1}{\tuple t_2}{\tuple t_3}$ if and only if for all $s, s' \in X$ with $\tuple t_1\langle s\rangle = \tuple t_1\langle s'\rangle$ there exists a $s'' \in X$ with $\tuple t_1 \tuple t_2\langle s''\rangle  = \tuple t_1 \tuple t_2\langle s\rangle $ and $\tuple t_1 \tuple t_3 \langle s''\rangle = \tuple t_1 \tuple t_3 \langle s'\rangle$.
\end{description}
In particular, it is not difficult to see that the above team satisfies $\indep{w_2}{w_1}{w_3}$, as required.\footnote{As observed in \cite{gradel13}, we then have that $=\!\!(x, y)$ is equivalent to $\indep{x}{y}{y}$, and so on. This can be justified along the same lines in which it was justified that $=\!\!(x)$ is equivalent to $\indepc{x}{x}$.}

It might be of course possible to consider other notions of dependence or independence, and justify them along similar lines. And indeed, in the opinion of the author at least, one of the main future research directions in the field of logics of imperfect information will be the search for new, doxastically significant notions of atom and the study of the relationships between the corresponding logics of imperfect information.

However, we already have more than enough basic material for our purposes.

Let us now add the classical \emph{conjunction} and \emph{disjunction} to our language:
\begin{description}
	\item[DI-or:] $M \models_X \psi \vee \theta$ if and only if $M \models_X \phi$ or $M \models_X \theta$.
	\item[DI-and:] $M \models_X \psi \wedge \theta$ if and only if $M \models_X \psi$ and $M \models_X \theta$.
\end{description}
There is not much to say here about these two connectives, as they are simply a way to join basic belief constraints into more complex ones.

Can we also add the classical quantifiers at this point? Nicely enough, this has been already done by Kontinen and V\"a\"an\"anen in \cite{kontinenv09} by defining the $\exists^1$ and $\forall^1$ quantifiers. These quantifiers have properties which are quite different from the ones of the ``usual'' existential and universal quantifiers for logics of imperfect information, which will be interpreted in Section \ref{sect:quantifiers} in terms of update operations.  Here we will simply write  $\exists$ and $\forall$ for the $\exists^1$ and $\forall^1$ of \cite{kontinenv09}. Their truth conditions are
\begin{description}
	\item[DI-exists:] $M \models_X \exists x \psi(x)$ if and only if there exists a $m \in \dom(M)$ such that $M \models_{X} \psi(m)$;
	\item[DI-forall:] $M \models_X \forall x \psi(x)$ if and only if for all $m \in \dom(M)$, $M \models_X \psi(m)$
\end{description} 
where $\psi(m)$ stands for the formula obtained by substituting $m$ (or, to be more precise, a new constant whose interpretation is the element $m$) for $x$ in $\psi(x)$.

These new connectives actually allow us to do without our many non first-order atoms: for example, it is not difficult to see at this point that $=\!\!(x, y)$ is equivalent to $\forall u \exists v \believes(u\not = x \vee v = y)$, $x \subseteq y$ is equivalent to $\forall u (\believes(u \not = x) \vee \possible(u=y))$, $x ~|~ y$ is equivalent to $\forall u(\believes(u \not = x) \vee \believes(u \not = y))$ and $\indep{x}{y}{z}$ is equivalent to $\forall u_1 u_2 u_3(\believes(u_1 \not = x \wedge u_2 \not = y) \vee \believes(u_1 \not = x \wedge u_3 \not = z) \vee \believes(u_1 = x \wedge u_2 = y \wedge u_3 = z))$.

Examining logics of imperfect information in terms of these quantifiers, rather than in terms of some classes of non first-order atom, may actually be a promising -- and, at the moment, largely unexplored -- avenue of research; but in this work, we thought it better to begin by describing the atoms that have been studied so far, and only show their translations in term of basic quantifiers in a second time.

Another connective that we are missing is a negation. We can certainly negate a first-order formula inside a belief or possibility statement; but can we also consider an ``external'' negation, to go with our external quantifiers and our classical conjunction and disjunctions?

It turns out that we can, of course, and that such an operator is precisely the \emph{contradictory negation} $\sim \phi$ of Team Semantics. Here we call it simply $\lnot \phi$, as there is no ``other'' negation with which it may be confused. Its semantics is precisely the one that we would expect: 
\begin{description}
	\item[DI-not:] $M \models_X \lnot \psi$ if and only if $M \not \models_X \psi$. 
\end{description} 
Given such an operator, we can as usual rewrite $\phi \wedge \psi$ as $\lnot(\lnot\phi \vee \lnot\psi)$ and $\forall x \psi$ as $\lnot \exists x \lnot \psi$.

Furthermore, as in the case of the ``diamond'' and ``box'' operators of Modal Logic, we can remove one of them from our list of primitives: for example, we could keep only $\possible(\phi)$, and define $\believes(\phi)$ as $\lnot \possible(\lnot \phi)$. Here, however, it must be noted that the roles of the internal and external negations are quite different: the former states that something is not true with respect to the whole team, while the negation of $\lnot \phi$ states that $\phi$ is not true \emph{in one specific assignment}. This is rather reminiscent, although not entirely identical, to the distinction between \emph{contradictory negation} and \emph{dual negation} of Team Logic.\footnote{In particular, the Team Logic expression $\sim \phi$ corresponds to our $\lnot \believes(\phi)$, while the expression $\lnot \phi$ corresponds to our $\believes(\lnot \phi)$.}

Given such a theory $T$ in the language developed so far and a suitable model $M$, one can consider the set $\bel_M(T)$ of all belief sets which satisfy $T$ in $M$: formally, we can define
\[
	\bel_M(T) := \{X : M \models_X T\}.
\]

Given two theories $T_1$ and $T_2$ and a suitable model $M$, we write $T_1 \models_M T_2$ if $\bel_M(T_1) \subseteq \bel_M(T_2)$, and $T_1 \models T_2$ if $T_1 \models_M T_2$ for all suitable models $M$.

The significance of these expressions is clear: $T_1 \models_M T_2$ means that whenever our belief set over the model $M$ satisfies all the conditions of $T_1$, it also describes all those described in $T_2$, and $T_1 \models T_2$ means that this is the case even if we do not know the underlying model $M$.
\section{Belief Updates}
\label{sect:updates}
Consider again our agent $A$ with his belief set $X_A$, and suppose that he interacts with another agent $B$ with a different belief set $X_B$. Then, our first agent could update his beliefs in many different ways, according to the degree up to which he trusts his own beliefs, to the degree up to which he trust the other agent's beliefs and on a number of other possible factors. 

This can be represented by defining update operations $X_A \diamond X_B$ from pairs of teams to \emph{sets} of teams; and, as stated in Section \ref{sect:belief_models}., we will write $X_A \diamond X_B \mapsto Y$ for $Y \in (X_A \diamond X_B)$, that is, for the assertion according to which $Y$ is a possible outcome of a $\diamond$-interaction between two agents whose beliefs are represented by $X_A$ and $X_B$ respectively.

There are many possible choices of update operations of this sort. Here we will consider four of them which seem, at least at first sight, to be relatively reasonable choices:
\begin{description}
	\item[Confident update $X_A \oplus X_B$:] $A$ trusts his beliefs concerning the true assignment $s_0$, but learns and trusts in the same way the beliefs of $B$ too. Therefore, $X_A \oplus X_B \mapsto Y$ if and only if $Y = X_A \cap X_B$. 
	\item[Credulous update $X_A \otimes X_B$:] $A$ is willing to entertain the possibility that he is wrong and the true state is one that $B$ believes possible and he does not. Or $B$ may be wrong and he may be right, he does not know. Hence, $X_A \otimes X_B \mapsto Y$ if and only if $Y = X_A \cup X_B$.
	\item[Skeptical update $X_A \ominus X_B$:] $A$ \emph{might} trust $B$'s beliefs, but only if $B$ appears to know more than $A$ -- that is, if $B$ does not consider possible anything that $A$ considers impossible. Otherwise, $A$ refuses to perform the update. Therefore, $X_A \ominus X_B \mapsto Y$ if and only if $X_B \subseteq X_A$ and $Y = X_B$.
	\item[Openminded update $X_A \odot X_B$:] $A$ might trust $B$'s beliefs, but only if $B$ appears to know \emph{less} than $A$ -- that is, if he does not consider impossible anything that $A$ considers possible. Otherwise, $A$ refuses to perform the update. Therefore, $X_A \odot X_B \mapsto Y$ if and only if $X_A \subseteq X_B$ and $Y = X_B$.
\end{description}
Note that the skeptical update and the openminded one can \emph{fail}, that is, may not lead to any possible outcome. This is a feature, not a bug: an update operation does not need to be specified for all possible belief states, and does not need to be deterministic either -- none of the update operations considered here can lead to more than one possible outcome, but nothing prevents in principle the definition of such update operators.

Furthermore, these update operators satisfy the three following properties:
\begin{description}
	\item[Idempotence:] $X \diamond X \mapsto Y$ if and only if $Y = X$;
	\item[Associativity:] If $X_1 \diamond X_2 \mapsto Y$ and $Y \diamond X_3 \mapsto Z$, then there exists a $W$ such that $X_2 \diamond X_3 \mapsto W$ and $X_1 \diamond W \mapsto Z$;
	\item[Monotonicity:] If $X \diamond Y \mapsto Z$ and $Z \diamond W \mapsto X$ then $X = Z$.
\end{description}
The interpretation of these properties, and the reason why they may be reasonable properties to require for an update operator, should be clear. An update $X \diamond Y$ represents an interaction between two agents whose beliefs correspond to $X$ and $Y$: therefore, idempotence states that whenever the two agents have the exact same beliefs, the interaction does not modify these beliefs, associativity states that, in group interactions, the order of the individual interactions is irrelevant, and monotonicity means that an agent who changed idea cannot ``return back'' to his previous beliefs through another interaction of the same kind. 

Let us verify that the update operators that we defined satisfy these properties. For the confident and credulous updates, this is obvious; hence, we will verify the case of the skeptical update, as the one of the openminded update is completely analogous.
\begin{description}
	\item[Idempotence:] Since $X \subseteq X$, it follows at once that $X \ominus X \mapsto Y$ if and only if $X = Y$.
	\item[Associativity:] Suppose that $X_1 \ominus X_2 \mapsto Y$ and $Y \ominus X_3 \mapsto Z$. Then, by the definition of the skeptical update, we have that $X_1 \supseteq X_2 = Y \supseteq X_3 = Z$. But then we have that $X_2 \ominus X_3 \mapsto X_3$, and that $X_1 \ominus X_3 \mapsto X_3 = Z$, as required.
	\item[Monotonicity:] If $X \ominus Y \mapsto Z$, then $Z = Y \subseteq X$. Furthermore, if $Z \ominus W \mapsto X$ then $X = W \subseteq Z$. Hence, $Z \subseteq X \subseteq Z$, and therefore $X = Z$.
\end{description}

Of course, these are only a possible selection of update properties: it may well be the case that in the future interesting update operators which do not respect them will be found, or that other, more important conditions will be explored.

After this parenthesis, let us reconsider the language of Section \ref{sect:atoms}. As we saw, this language is already capable of describing a number of properties of belief states. But now we have something new, that is, some ways of \emph{updating} beliefs. This allows us to ask some very natural questions: for example, if $A$'s belief state $X_A$ satisfies $\phi$, and if $B$'s belief state $X_B$ satisfies $\psi$, what can we say about a belief state corresponding to some update of $X_A$ and $X_B$?

To answer this, we need to add some way of talking about updates to our language. A possibility is to consider, for each update $\diamond$, a connective $\phi \diamond \psi$ such that $M \models_X \phi \diamond \psi$ if and only if the belief set $X$ can be seen as the result of a $\diamond$-update between a belief set satisfying $\phi$ and a belief set satisfying $\psi$. 

In other words, the semantics of $\phi \diamond \psi$ will be
\begin{description}
\item[DI-$\diamond$:] $M \models_X \phi \diamond \psi$ if and only if there exist teams $Y$ and $Z$ such that\\$Y \diamond Z \mapsto X$, $M \models_Y \phi$ and $M \models_Z \psi$.
\end{description}

This gives us at once the following operations: 
\begin{description}
	\item[DI-$\oplus$:] $M \models_X \phi \oplus \psi$ if and only if $X = Y \cap Z$ for some $Y$ and $Z$ such that $M \models_Y \phi$ and $M \models_Z \psi$; 
	\item[DI-$\otimes$:] $M \models_X \phi \otimes \psi$ if and only if $X = Y \cup Z$ for some $Y$ and $Z$ such that $M \models_Y \phi$ and $M \models_Z \psi$;
	\item[DI-$\ominus$:] $M \models_X \phi \ominus \psi$ if and only if $M \models_X \psi$ and there exists a $Y \supseteq X$ such that $M \models_Y \phi$.
	\item[DI-$\odot$:] $M \models_X \phi \odot \psi$ if and only if $M \models_X \psi$ and there exists a $Y \subseteq X$ such that $M \models_X \phi$.
\end{description}
The credulous update connective is exactly the tensor connective of Team Logic, or the disjunction of Dependence Logic. The other ones, to the knowledge of the author, have not been studied in depth yet, but they do hold some interest; it is worth noting, in particular, that for downwards closed logics (such as Dependence Logic or Intuitionistic Dependence Logic) $\phi \oplus \psi$ and $\phi \ominus \psi$ are equivalent and correspond to the classical conjunction $\phi \wedge \psi$.

The intended interpretation of these connectives is better understood by considering expressions of the form $\phi \diamond \psi \models \theta$. According to what we just discussed, such an expression corresponds to the statement that  
\begin{quote}
	\emph{Any possible outcome of $\diamond$-update between two belief states satisfying $\phi$ and $\psi$ respectively will satisfy $\theta$.}
\end{quote}
The significance of such a statement, and its relevance for the kind of framework that we are presenting, is then clear.

An expression of the form $\phi \models \psi \diamond \theta$ is perhaps a little more difficult to read, but its meaning is still intuitive enough: such an entailment holds if and only if any belief state which satisfies $\phi$ can be thought of as the result of a $\diamond$-update between a belief state such that $\psi$ and a belief state such that $\theta$. 

This allows us to make sense of some properties of logics of imperfect information. Here we will describe only three examples relative to the credulous update: 
\begin{enumerate}
	\item It is easy to see that $(\phi \otimes \psi) \otimes \theta \models \phi \otimes (\psi \otimes \theta)$ for all $\phi$, $\psi$ and $\theta$. This just means that the credulous update is \emph{associative}: if a belief state can be the result of an agent, whose belief state satisfies $\phi$, performing a credulous update with some agent whose belief set satisfies $\psi$ and \emph{then} with some other one whose belief set satisfies $\theta$ , then it can also be the result of an agent, whose belief state satisfies $\phi$, performing a credulous update with some agent whose belief state \emph{satisfied} $\psi$ before \emph{he} performed a credulous update with some agent whose belief state satisfied $\theta$.
	\item The credulous update is not idempotent: in general, $\phi \otimes \phi \not \models \phi$. This can be verified easily by letting $\phi$ be the constancy atom $=\!\!(x)$: if the value of the variable $x$ is constant in $Y$ and in $Z$, indeed, it does not necessarily follow that it is constant in $Y \cup Z$.

		The reason for this is clear: if an agent who believes that he knows the value of $x$ performs a credulous update with another agent who also believes that he knows the value of $x$, and the two agents disagree on this value, then our first agent will become unsure about who, between him and the other agent, was in the right about $x$. 
		
	\item If $\phi$, $\psi$ and $\theta$ are downwards closed formulas -- for example, if they are expressible in Intuitionistic Dependence Logic or in Exclusion Logic -- then the following ``distributivity property'', first pointed out by Ville Nurmi, holds:
\[
	(\phi \otimes \psi) \wedge (\phi \otimes \theta) \models \phi \otimes \phi \otimes (\psi \wedge \theta).
\]
According to what we just discussed, this entailment can be read as follows: 
\begin{quote}
	\emph{If a team $X$ can be seen as the result of a credulous update between a team such that $\phi$ and one such that $\psi$, and also as the result of a credulous update between a team such that $\phi$ and one such that $\theta$, then it is can also be the result of a credulous update between two teams such that $\phi$ and one such that $\psi$ and $\theta$.}
\end{quote}
This is a nontrivial -- and, at least in the opinion of the author, rather interesting -- property concerning belief updates and their properties. 
\end{enumerate}
As an aside, the last property fails if we consider non-downwards closed formulas: for example, consider the team 
\[
X = \begin{array}{c | c c c}
	& x & y & z\\
	\hline
	s_0 & 0 & 1 & 1\\
	s_1 & 1 & 0 & 0
\end{array}
\]
in any model $M$ with at least two elements. Then $M \models_X z = 1 \otimes x \subseteq y$: indeed, for $Y = \emptyset$ and $Z = X$ we have that $M \models_Y x=1$, $M \models_Z x \subseteq y$, and $X = Y \cup Z$. Furthermore, $M \models_X z=1 \otimes z=0$, as can be easily verified by splitting $X$ into the two subteams $\{s_0\}$ and $\{s_1\}$.

However, $M \not \models_X z=1 \otimes z=1 \otimes (x \subseteq y \wedge z=0)$: indeed, otherwise we could split $X$ into three subteams $X_1$, $X_2$ and $X_3$ such that $M \models_{X_1} z=1$, $M \models_{X_2} z=1$ and $M \models_{X_3} x \subseteq y \wedge z = 0$. Now, $s_1(z) = 0 \not = 1$, and therefore $s_1$ would necessarily be in $X_3$; but then, since $M \models_{X_3} x \subseteq y$, there should be another assignment $s \in X_3$ with $s(y) = s_1(x) = 1$. The only such assignment is $s_0$; but $s_0(z) = 1$, and therefore it would not be the case that $M \models_{X_3} z = 0$. This contradicts our hypothesis.

As this example shows, different fragments of our language may have different properties when it comes to the entailment relation. This may be worth exploring further in the future.
\section{Adjoints}
\label{sect:adjoints}
In the previous section we considered a few update operators, and for each one of them we defined a connective expressing that a given team $X$ can be seen as the result of an update between some teams $Y$ and $Z$ satisfying certain properties. This increased substantially the expressive power of our formalism, and, in fact, by now our language contains the propositional fragment of most logics of imperfect information.

However, this is not all that we can do with these update operations. In particular, it may be useful to be able to make conjectures about what \emph{would} happen if we updated the team in a certain way. In particular, any update operator $\diamond$ induces a corresponding implication $\xrightarrow{\diamond}$ between sets of belief sets, defined as 
\begin{description}
\item[DI-$\xrightarrow{\diamond}$:] $M \models_X \phi \xrightarrow{\diamond} \psi$ if and only if for all $Y$ s.t. $M \models_Y \phi$ and for all $Z$  such that $X \diamond Y \mapsto Z$, $M \models_{Z} \psi$.
\end{description}
The intended interpretation of these new connectives is clear: $\phi \xrightarrow{\diamond} \psi$ corresponds to the statement asserting that \emph{any} $\diamond$-update between the beliefs $X$ of our agent and the beliefs $Y$ of \emph{any} other agent such that $M \models_Y \phi$ will \emph{always} result in a belief $Z$ such that $\psi$.

This easily implies that
\[
	\phi \diamond \psi \models_M \theta \Leftrightarrow \phi \models_M \psi \xrightarrow{\diamond} \theta
\]
for all $\phi$, $\psi$ and $\theta$ and for all models $M$. In other words, the operator $\xrightarrow{\diamond}$ is the \emph{right adjoint} of the operator $\diamond$.

This notion of \emph{adjointness} is precisely the one studied in \cite{abramsky09}, in which it was one of the motivations given for the definitions of the intuitionistic and linear implications.

Hence, it should come to no surprise that now our framework will allow us to recover both these implications, plus two new ones. Indeed, by instantiating our definition of $\xrightarrow{\diamond}$ with the update operators of the previous section we obtain the following connectives:
\begin{description}
	\item[Confident implication:] $M \models_X \phi \confimp \psi$ if and only if for all $Y$ such that $M \models_Y \phi$ it holds that $M \models_{X \cap Y} \psi$;
	\item[Credulous implication:] $M \models_X \phi \credimp \psi$ if and only if for all $Y$ such that $M \models_Y \phi$ it holds that $M \models_{X \cup Y} \psi$;
	\item[Skeptical implication:] $M \models_X \phi \skepimp \psi$ if and only if for all $Y \subseteq X$ such that $M \models_Y \psi$ it holds that $M \models_Y \theta$;
	\item[Openminded implication:] $M \models_X \phi \openimp \psi$ if and only if for all $Y \supseteq X$ such that $M \models_Y \psi$ it holds that $M \models_Y \theta$. 
\end{description}
Skeptical implication and credulous implication are precisely the intuitionistic and linear implications of \cite{abramsky09}. Moreover, it is not difficult to see that whenever the antecedent satisfies the downwards closure property, the confident and the skeptical implications are equivalent (as would the credulous and the openminded ones in the case of an upwards closed antecedent). This, in particular, implies that for downwards closed logics (such as, for example, Intuitionistic Dependence Logic) these two forms of implication are interchangeable.

However, in general the skeptical and the confident implications are not equivalent. For example, consider the team 
\[
	X = \begin{array}{c | c c}
		& x & y\\
		\hline
		s_0 & 0 & 0\\
		s_1 & 1 & 1
	\end{array}
\]
in a model $M$ with two elements $0$ and $1$. Then $M \models_X \indepc{x}{y} \skepimp =\!\!(x)$: indeed, the only subteams of $X$ in which $x$ is independent of $y$ are $\{s_0\}$ and $\{s_1\}$, and in these subteams $x$ is clearly constant. 

But $M \not \models_X \indepc{x}{y} \confimp =\!\!(x)$: indeed, for
\[
	Y = \begin{array}{c | c c}
		& x & y\\
		\hline
		s_0 & 0 & 0\\
		s_1 & 1 & 1\\
		s_2 & 0 & 1\\
		s_3 & 1 & 0
	\end{array}
\]
we have that $M \models_Y \indepc{x}{y}$, but that in $X \cap Y = X$ the value of $x$ is not constant.

So now we have a new class of formulas which describe beliefs in terms of how they \emph{would} change if they interacted with other beliefs; and this is, of course, of potential significance for a number of practical applications. More in general, it seems that the problem of deciding, given two formulas $\phi$ and $\psi$ of our language (or of a fragment thereof) and a fixed model $M$, whether $\phi \models_M \psi$, is of no small relevance for the field of knowledge updating; and that the same may also be said for the problem of whether $\phi$ entails $\psi$ in all models.

The interpretation just discussed also clarifies the fact, already pointed out in \cite{abramsky09}, that $=\!\!(x, y)$ is logically equivalent to $=\!\!(x) \skepimp =\!\!(y)$ (or equivalently, since constancy atoms are downwards closed, to $=\!\!(x) \confimp =\!\!(y)$), and that more in general dependence atoms can be decomposed in terms of constancy atoms and intuitionistic implication: indeed, a belief set $X$ satisfies $=\!\!(x) \confimp =\!\!(y)$ if and only if the corresponding agent, by trusting the beliefs $Y$ of some agent who believes he knows the value of $x$, will reach a new belief state $X \cap Y$ in which he believes to know the value of $y$ too. This corresponds precisely to the doxastic interpretation of $=\!\!(x, y)$. 
%
%
%
\section{Minimal updates}
\label{sect:minimality}
Let $\diamond$ be an update operator satisfying the \emph{idempotence}, \emph{associativity} and \emph{monotonicity} conditions described in Section \ref{sect:updates}.
Then $\diamond$ defines a \emph{partial order} over belief sets as follows:
\[
X \leq^{\diamond} Y \Leftrightarrow \exists X' \mbox{ s.t. } X \diamond X' = Y.
\]
Indeed, by idempotence we have that $X \leq^{\diamond} X$; by the associativity of the operator, we have that the $\leq^{\diamond}$ relation is transitive; and by the monotonicity of the operator, we have that if $X \leq^{\diamond} Y$ and $Y \leq^{\diamond} X$ then $X = Y$. 

The interpretation of the $\leq^{\diamond}$ operator in our framework is the following: $X \leq^{\diamond} Y$ if and only if an agent, whose belief set is $X$, \emph{may} reach the belief state $Y$ through a sequence of $\diamond$-updates.

Different update operators, of course, may generate the same partial order. In particular, for the operators that we considered we have that 
\[
	X \leq^{\oplus} Y \Leftrightarrow X \leq^{\ominus} Y \Leftrightarrow Y \subseteq X
\]
and
\[
	X \leq^{\otimes} Y \Leftrightarrow X \leq^{\odot} Y \Leftrightarrow X \subseteq Y.
\]
In other words, a belief state $Y$ can be reached from a state $X$ through a confident or a skeptical update if and only if $Y$ represents a \emph{stricter} belief than $X$ does, and it can be reached through a credulous or an openminded statement if and only if it represents a \emph{looser} belief than $X$ does. 

By the way, this allows us to give an alternative definition of the skeptical and openminded updates in terms of the confident and credulous ones as follows:
\begin{description}
	\item[Skeptical update, v2:] $X \ominus Y \mapsto Z$ if and only if $X \leq^{\oplus} Y$ and $X \oplus Y \mapsto Z$;
	\item[Openminded update, v2:] $X \odot Y \mapsto Z$ if and only if $X \leq^{\otimes} Y$ and $X \otimes Y \mapsto Z$.
\end{description}
This seems to be an instance of a more general phenomenon: given an update operation $\diamond$ satisfying our three conditions, we can always generate a new operation $\diamond'$ as 
\[
	X \diamond' Y \mapsto Z \Leftrightarrow X \leq^{\diamond} Y \mbox{ and } X \diamond Y = Z.
\]
These new update operations $\diamond'$, in other words, are defined precisely as the older operations $\diamond$, except that now our agent -- who believes that $X$ -- is willing to perform an update with $Y$ if and only if $Y$ itself is a belief state that he \emph{could} possibly reach through a $\diamond$-update.

Now, let $\mathcal V$ be any family of belief sets, let $X$ be a belief, and let us define $X \diamond \mathcal V$ as $\{Z : \exists Y \in \mathcal V \mbox{ s.t. } X \diamond Y \mapsto Z\}$. As usual, we will write $X \diamond \mathcal V \mapsto Z$ for $Z \in (X \diamond \mathcal V)$: in other words, with $X \diamond \mathcal V \mapsto Z$ we mean that $Z$ is a possible outcome of updating $X$ with some $Y \in \mathcal V$.

Suppose now that our agent can choose which $Y \in \mathcal V$ to pick to update his beliefs, and also select the resulting $Z$ if more than one exists: which strategy could he use?

A reasonable choice might be that our agent will attempt to make a $\diamond$\emph{-minimal} update, that is, one that does not commit him any more than necessary: in particular, if he can reach both $Z_1$ and $Z_2$, and he could reach $Z_2$ from $Z_1$ through \emph{another} $\diamond$-update, then he should pick $Z_1$ over $Z_2$. This can be defined formally as the notion of \emph{minimal update}:
\begin{description}
\item $X ~\Box~ \mathcal V \mapsto Z$ if and only if there is a $Y$ such that $X \diamond Y \mapsto Z$ and $Z$ is $\leq^{\diamond}$-minimal in $X \diamond \mathcal V$.
\end{description}
Here, stating that $Z$ is $\leq^{\diamond}$-minimal in $X \diamond \mathcal V$ means simply that there exists no $Z' \in X \diamond \mathcal V$ with $Z' <^{\diamond} Z$.

Substituting $\diamond$ with the four update operators considered so far, we get the following updates: 
\begin{description}
		\item[Minimal confident update:] $X \boxplus \mathcal V \mapsto Z$ if and only if there exists a $Y \in \mathcal V$ such that $X \cap Y = Z$, and if for all $Z' \supsetneq Z$ and all $Y' \in \mathcal V$ it holds that $X \cap Y' \not = Z'$;
	\item[Minimal credulous update] $X \boxtimes \mathcal V \mapsto Z$ if and only if there exists a $Y \in \mathcal V$ such that $X \cup Y = Z$, and for all $Z' \subsetneq Z$ and all $Y' \in \mathcal V$ it holds that $X \cup Y' \not = Z'$;
	\item[Minimal skeptical update:] $X \boxminus \mathcal V \mapsto Z$ if and only $Z \subseteq X$, $Z \in \mathcal V$ and for all $Z'$ with $Z \subsetneq Z' \subseteq X$ it holds that $Z' \not \in \mathcal V$;
	\item[Minimal openminded update:] $X \boxdot \mathcal V \mapsto Z$ if and only if $X \subseteq Z$, $Z \in \mathcal V$, and for all $Z'$ with $X \subseteq Z' \subsetneq Z$ it holds that $Z' \not \in \mathcal V$.
\end{description}
As before, we can at this point define connectives $\phi ~\Box~\psi$ for describing that a team $X$ is a possible result of a minimal update of this kind, and connectives $\phi \xrightarrow{\Box} \psi$ for describing that whenever we perform a minimal update between $X$ and a the family of teams satisfying $\phi$, the result will satisfy $\psi$.

 The formal definitions would then be
\begin{description}
\item[DI-$\Box$:] $M \models_X \phi ~\Box~ \psi$ if and only if there exists a $Y$ such that $M \models_Y \phi$ and $Y ~\Box~ \bel_M(\psi) \mapsto X$;
\item[DI-$\xrightarrow{\Box}$:] $M \models_X \phi \xrightarrow{\Box} \psi$ if and only if whenever $X ~\Box~ \bel_M(\phi) \mapsto Z$ it holds that $M \models_Z \psi$.
\end{description}
Here, as usual, $\bel_M(\psi)$ represents the family of all teams which satisfy $\psi$ in $M$.

Here we will not give the instantiations of these connectives for the four updates described above, nor will we discuss their properties.

All that we will point out is that the minimal skeptical implication connective $\phi \xrightarrow{\boxminus} \psi$ is precisely the  \emph{maximal implication} $\phi \hookrightarrow \psi$ mentioned in \cite{kontinennu09} and defined as
\begin{description}
	\item[DI-maximp:] $M \models_X \phi \hookrightarrow \psi$ if and only if for all $Y \subseteq X$ such that $M \models_Y \phi$ and $M \not \models_Z \phi$ for all $Z$ with $Y \subsetneq Z \subseteq X$, $M \models_Y \psi$; 
\end{description}

Thus, even this connective can be interpreted in this framework. This notion of minimal update appears to be rather natural, and it probably deserves further study; however, for the moment we will content ourselves with having defined it and shown how to recover the $\hookrightarrow$ implication through it.

As an aside, this implication allows us to decompose independence atoms: for example, it is not difficult to see that $\indep{x}{y}{z}$ is equivalent to $=\!\!(x) \hookrightarrow \indepc{y}{z}$.
\section{Quantifiers}
\label{sect:quantifiers}
So far, our operators have treated assignments as if they were point-like possible worlds. This is not the case, of course: for example, our agent may be confident about the values of certain variables, but not about the ones of others. Furthermore, we have no way so far of adding or removing variables to the domains of our teams. In this section, we will attempt to remedy this.

Let us begin with a \emph{forgetting} operator $\rho x$, where $x$ is a variable, with has the effect of removing the variable $x$ from the domain of our team: more precisely, for all belief states $X$ we define $\rho x (X)$ as $X_{\backslash x}$, that is, as the team containing the restrictions of all assignments in $X$ to $\dom(X) \backslash \{x\}$. In the case that $x$ is not in the domain of $X$ to begin with, this operator has no effect.

The doxastic meaning of this operator is the one suggested by its value: after performing the update $(\rho x)$, our agents forgets everything about the value of the variable $x$, and even the fact that this variable exists to begin with! This is not the same as our agent simply professing ignorance about the value of $x$ -- this would be another operator, that we will examine later -- as here we are \emph{really} erasing the variable from our domain.

As in the case of the binary update operators considered in the previous sections, this forgetting operator corresponds to \emph{two} distinct connectives, which can be formally defined as 
\begin{description}
	\item[DI-forgotten:]$M \models_X (\rho x) \phi$ if and only if there exists a team $X'$ such that $X = (\rho x)X' = X'_{\backslash x}$ and $M \models_{X'} \phi$. 
	\item[DI-forgetting:] $M \models_X (\eta x) \phi$ if and only if $M \models_{X'} \phi$, where $X' = (\rho x)X = X_{\backslash x}$;
\end{description}
In other words, $(\eta x) \phi$ holds in a team $X$ if, starting from $X$ and forgetting the values of the variable $x$, we obtain a belief state which satisfies $\phi$, and $(\rho x)\phi$ holds in a team $X$ if this team can be obtained by starting from a team $X'$ which satisfies $\phi$ and forgetting the value of $x$. As always, these two operators are adjoints, that is, 
\[
	(\rho x) \phi \models \psi \Leftrightarrow \phi \models (\eta x) \psi.
\]

A combination of these two operators which is of particular importance is the \emph{disbelieving operator} $\disbelieve x \phi =  (\eta x)(\rho x) \phi$. The intuition here is that $M \models_X \disbelieve x \phi$ if, apart from the value of the variable $x$, the team $X$ could correspond to a belief set which satisfies $\phi$. This is evident by the corresponding semantic rule: indeed, by combining the rules for the forgetting and remembering operators, one can see that $M \models_X \disbelieve x \phi$ if and only if there exists a $X'$ such that $X_{\backslash x} = X'_{\backslash x}$ and such that $M \models_{X'} \phi$. 

The condition $X_{\backslash x} = X'_{\backslash x}$ is easily seen to be equivalent to the existence of a function $H: X \rightarrow \mathcal P(\dom(X))\backslash \{\emptyset\}$ such that $X' = X[H/x] = \{s[m/x] : s \in X, m \in H(s)\}$. Therefore, the $\disbelieve x \phi$ operator corresponds precisely to the existential quantifier rule.\footnote{To be more precise, it corresponds to the \emph{lax} variant of the existential quantifier discussed in \cite{galliani12}, in which more than one possible value for $x$ may be picked for each assignment.}

Just like all connectives, $\disbelieve x \phi$ can also be built directly from some belief update operator. This is our first true case of a \emph{non-deterministic} belief update: more precisely, we can define it as 
\[
(\disbelieve x) X \mapsto Y \mbox{ if and only if } X_{\backslash x} = Y_{\backslash x}.
\]
The significance of this operator in our framework should be easy to see: in brief, we have that $(\disbelieve x) X \mapsto Y$ if and only if an agent who starts from the belief $X$ and, disbelieving his previous opinions about the possible values of the variable $x$, changes them \emph{and nothing else}, can possibly reach the belief state represented by $Y$.

The quantifier $(\disbelieve x) \phi$ is then the unary equivalent of the $\phi \diamond \psi$ connectives considered in the previous sections: in brief, $M \models_X (\disbelieve x) \phi$ if and only if there exists a team $Y$ with $M \models_Y \phi$ and $(\disbelieve x) Y \mapsto X$.

Of course, we also get another quantifier $(\regardless x) \phi$, which is satisfied by a belief state $X$ if and only if for all $Y$ such that $(\disbelieve x) X \mapsto Y$ we have that $M \models_Y \phi$: in other words, $M \models_X (\regardless x) \phi$ corresponds that our agent, whose belief state is $X$, is confident that $\phi$ would hold even if his beliefs about the value of $x$ were wrong. Hence, we may perhaps call it the \emph{regardless} quantifier. Thus, we have obtained a new pair of unary connectives: 
\begin{description}
	\item[DI-disbelief:] $M \models_X (\disbelieve x) \phi$ if and only if there exists a $Y$ such that $Y_{\backslash x} = X_{\backslash x}$ and $M \models_{Y} \phi$;
	\item[DI-regardless:]$M \models_X (\regardless x) \phi$ if and only if for all teams $Y$ with $Y_{\backslash x} = X_{\backslash x}$ it holds that $M \models_Y \phi$.
\end{description}
The ``regardless'' operator is a lax version of the universal quantifier $\sim \exists x \sim \ldots$ of Team Semantics; and, as always, these two operators are adjoints, that is, 
\[
	(\disbelieve x) \phi \models \psi \Leftrightarrow \phi \models (\regardless x) \psi.
\]

Finally, let us consider the following scenario: our agent's belief is represented by the team $X$, which satisfies some property $\phi$, but now the agent decides that he does not trust at all his own opinion about the value of some variable $x$. Then the new belief state is given by 
\[
	\doubted x \phi = X[M/x] = \{s[m/x] : s \in X, m \in \dom(M)\}
\]
that is, the new belief state of our agent is the same as the old one, except that now our agent knows nothing at all about $x$. This, once again, represents a situation in which our agent doubts the validity of his beliefs about $x$; but where the disbelief operator $\disbelieve x$ corresponds to the agent revising these beliefs in some arbitrary, nondeterministic way, this new operator $\doubted x$ has the agent taking an agnostic position about the possible values of $x$ and moving to a belief state in which he knows nothing about it.

As in all previous cases, this allows us to develop two new connectives, that we will call the \emph{doubted} and the \emph{doubting} quantifiers:
\begin{description}
	\item[DI-doubted:] $M \models_X (\doubted x) \phi$ if and only if $X = Y[M/x]$ for some $Y$ such that $M \models_Y \phi$.
	\item[DI-doubting:] $M \models_X (\doubting x) \phi$ if and only if $M \models_{X[M/x]} \phi$;
\end{description}
The $\doubting x$ connective is exactly the $!x$ operator in Team Logic, or the one written as $\forall x$ in Dependence Logic or in many other logics of imperfect information. The other one is, to the knowledge of the author, new, but its interpretation is clear: $M \models_X (\doubted x) \phi$ if the belief $X$ can be seen as the result of taking a belief state $Y$ which satisfies $\phi$, and doubting its guess about it. As always, we have that 
\[
	\doubted x \phi \models \psi \Leftrightarrow \phi \models \doubting x \psi.
\]
\section{Conclusion}
As we saw, virtually all of the connectives and operators of Dependence Logic and its extensions admit natural interpretations in terms of descriptions of beliefs and/or belief update operations. This suggests that we can extend our formalism along at least two different lines, by adding more \emph{dependency notions} for the specification of more properties of beliefs or by adding further forms of belief update. It will be interesting to examine how these two aspects will interact. 

We mention some possible avenues for further research: 
\begin{enumerate}
\item Modal Logics are, \emph{de facto}, the standard logical formalisms for reasoning about beliefs; and in particular, \emph{dynamic modal logics} \cite{baltag98,plaza07,vanbenthem07} add to the basic modal framework the capability for reasoning about belief updates. It is high time for a more formal examination of the relationship between these approaches and Team Semantics, perhaps along the lines of a study of the doxastic content of Modal Dependence Logic \cite{vaananen08b} and its extensions. The results promise to be greatly rewarding for both the Dependence Logic and the Dynamic Modal Logic research communities.
\item Likewise, Belief Revision concepts and methodologies can (and, in the opinion of the author, \emph{should}) be adapted to Team Semantics. We only dealt with very simple notions of belief update so far; but there is in principle no reason why the more complex ones which have been studied within the Belief Revision research area could not be introduced in Team Semantics. Our notion of \emph{update minimality}, in particular, appears to have some connections with the \emph{distance semantics for belief revision} of \cite{lehmann01}.
\item If teams are to represent belief states as set of possible assignments, and if team semantics can be thought of as a formalism for reasoning about these states and their evolution, it follows that it should be possible to develop analogous logical formalisms for finer-grained forms of belief representation. Just to mention one possibility, it may be very worthwhile to consider \emph{probabilistic distributions} over assignments, that is, \emph{probabilistic teams} in the sense of \cite{galliani08}. This might bring about interesting connections to the \emph{Equilibrium Semantics} of \cite{sevenster09,gallmann10}, but from a radically different perspective: rather than searching for the values of the strategic equilibria of the semantic games corresponding to sentences, as is the case for Equilibrium Semantics, we would associate formulas to conditions over probability distributions over assignments.
\end{enumerate}
As commented by a reviewer, many of the ideas and results of the work of this paper does not seem to depend intrinsically from the definition of teams as sets of assignments. A -- potentially very fruitful -- avenue of further research may consist in abstracting from this definition and study belief representation through \emph{algebraizations} of Team Semantics, based for example on Mann's work on IF Logic \cite{mann09}: this would increase further the level of generality and abstraction of our framework. 

We leave a more in-depth analysis of these ideas to future works. Here we limited ourselves to a discussion of the doxastic interpretation of the \emph{current} state of the art in Team Semantics research; and it is the hope of the author that this presentation highlighted how this interpretation holds much promise for the further development of this fascinating family of logics.
\section*{Acknowledgments:}
We thank Samson Abramsky, Alexandru Baltag. Johan van Benthem and Jouko V\"a\"an\"anen for many insightful comments.  We also thank an anonymous reviewer for their useful suggestions and corrections.
\bibliographystyle{plain}
\bibliography{biblio}
\end{document}